\documentclass{article}

\usepackage{spconf}
\usepackage[numbers,sort&compress]{natbib}

\usepackage{amsmath, amssymb, amsfonts, amsthm}
\usepackage{bm}

\usepackage{graphicx}
\usepackage{float}
\usepackage{tikz}
\usetikzlibrary{positioning}
\usetikzlibrary{arrows.meta}
\usetikzlibrary{shapes.geometric}
\usepackage{adjustbox}
\usepackage[font=small]{caption} % Adjust caption font
\usepackage{booktabs}

\usepackage{subcaption}

\title{\normalsize Monocular 3D Object Position Estimation with VLMs for Human-Robot Interaction}
\name{Ari Wahl$^1$$^2$, Dorian Gawlinski$^1$$^2$, David Przewozny$^1$, Paul Chojecki$^1$, Felix Bießmann$^2$, Sebastian Bosse$^1$
\thanks{This research was supported by the Governing Mayor of Berlin, Senate Chancellory Research and Technology (LZDV)}
}
\address{\small
    $^1$ Fraunhofer HHI, Berlin, Germany\\
    \small
    $^2$ Berliner Hochschule für Technik (BHT), Berlin, Germany\\
}

\begin{document}
\maketitle
\begin{abstract}Pre-trained general-purpose Vision-Language Models (VLM) hold the potential to enhance intuitive human-machine interactions due to their rich world knowledge and 2D object detection capabilities \cite{kawaharazuka2023}. 
However, VLMs for 3D coordinates detection tasks are rare. In this work, we investigate interactive abilities of VLMs by returning 3D object positions given a monocular RGB image from a wrist-mounted camera, natural language input, and robot states. We collected and curated a heterogeneous dataset of more than 100,000 images and finetuned a VLM using QLoRA with a custom regression head. By implementing conditional routing, our model maintains its ability to process general visual queries while adding specialized 3D position estimation capabilities.
% The gripper orientation is fixed to a downward angle to ensure a good view of the workspace. 
%The main contribution of this work is to achieve the task while preserving the ability of the base model. 
Our results demonstrate robust predictive performance with a median MAE of 13 mm on the test set and a five-fold improvement over a simpler baseline without finetuning. In about 25\% of the cases, predictions are within a range considered acceptable for the robot to interact with objects.
\end{abstract}


\section{Introduction}
Vision-Language Models have been demonstrated to be useful in robotic settings for some tasks, such as Visual Question-Answering or Visual Dialogue for robotic tasks \cite{Firoozi2024}. For other tasks in robotics the application of Vision-Language Models remains subject of active research as can be seen in the following. We aim to tackle a critical gap in VLMs for robotics by enabling 3D object position estimation from single monocular RGB images. The adaptive feature routing dynamically directs visual features through specialized pathways depending on the task: regressing 3D coordinates or preserving general VLM capabilities. With the utilization of Low-Rank Adaptation (LoRA) matrices, we keep the base model intact while training only the LoRA and regression head parameters during finetuning. This feature is later leveraged to implement a conditional routing mechanism to still be able to access the original quantized base model and to use its full capabilities. We achieve a final model size of 3.7 B parameters. On the testset our final model has a median MAE of 13 mm and a median Euclidean distance error of 27 mm.

\section{Related Work} Vision-Language Models (VLM) for robotics are an accelerating field for which state-of-the-art (SOTA) is constantly changing. The field is rapidly growing and uses a wide spectrum of approaches, for example, Vision-Language Action Models (VLA) that output actions directly or VLMs that are also trained for pointing to objects in the 2D image space.
Some of the most well-known models that use vision and language backbones for robotics are models of the RT-X family. 
This is a family of VLA, which directly output robot actions. The RT-1-X (2023) and RT-2-X (2023) models are based on their predecessors, the two model architectures RT-1\cite{brohan2023rt1} and RT-2\cite{brohan2023rt2}, and are
trained on the Open X-Embodiment dataset, which features a wide range of robotic settings and tasks. They generalize
much better than their predecessors but are still not general robotic models. For RT-1-X, which is open source, studies have been conducted showing that it needs additional fine-tuning to be used on a robot that has not been part of the training set\cite{salzer2024_rt1x_to_scara},
the same is the case for OpenVLA\cite{kim2024openvla}, CogACT (2024)\cite{li2024cogact} and \(\bm{\pi}\)0 (2024)\cite{black2024pi0}, other promising open-source VLA trained on the Open X-Embodiment dataset. 
CogACT (2024) uses a VLM for cognitive information to guide an action prediction with a specialized action module. The \(\bm{\pi}\)0 (2024) model uses a VLM base model and an action expert. LERF\cite{kerr2023lerflanguageembedded} models have also been used for robotics, the most well known are LERF-TOGO (2023)\cite{rashid2023lerf-togo} and LifelongLERF (2024)\cite{rashid2024lifelonglerf}, which has shown progress in the direction of real-time capabilities by only re-rendering changing elements in the scene. Lately, general-knowledge VLMs are increasingly also trained for pointing on objects in the 2D space. Examples are RoboPoint(2024)\cite{yuan2024robopoint}, Molmo (2024)\cite{deitke2024molmopixmo} and Moka (2024)\cite{liu2024moka}, which projects these points into 3D with the help of RGB-D depth information. Typically, VLMs for monocular 3D object detection, such as VFMM3D (2024)\cite{ding2024vfmm3d} and MonoDETR (2023)\cite{zhang2023monodetr} are trained on autonomous driving datasets (e.g. Kitti). Monocular RGB camera input is possible for the RT-X family, LERF-TOGO, VFMM3D, MonoDETR and Molmo. However, only OpenVLA, RT-1-X, LERF-TOGO and CogACT are open-source. Only a few models have general-purpose VLMs as base models that are still fully accessible, such as RoboPoint. Although the list is not exhaustive, models that fit all our criteria and can be used out-of-the-box are not open-source yet.

\section{Methods}
\subsection{System architecture and processing pipeline}
The model was trained using Apptainer on a Slurm cluster and five NVIDIA Tesla A100 GPUs with 40 GB VRAM for 4 days.
The central task discussed in this paper is the open set detection of an object's position in a robotic setting with the use of a Vision-Language Model (VLM) and a single monocular camera on a robot arm. This task implies detection of 3D coordinates of an object in the robot's workspace relative to the robot's base. The pre-trained Vision-Language Model's object detection abilities are leveraged to achieve this. Additionally, the idea is that the VLM can also provide useful affordance detection abilities for downstream tasks. In contrast to most other work, this work is focused on preserving general abilities of the Vision-Language Model and only adding the 3D coordinate estimation as an additional ability. For this, a conditional routing mechanism is used to route general questions to the base model, and the task-related questions to the adapted model's architecture. As proof of concept, an intentionally simple signifier, the word "question" is used to rout specific prompts solely to the base model. Input is a single RGB image and a text prompt, including the current gripper position and orientation relative to the base.

\subsection{Experimental setup}
Data scarcity is one of the main problems for further improvements in these fields, another is the high variability of tasks, environments, and robotic platforms \cite{Firoozi2024}. To finetune a model for our task, the data had to be collected from the workspace of the used robot arm.
We collected the training data with a 6-joint Doosan A0509 robot arm equipped with a RG2-FT gripper and a Logitech Brio Webcam mounted on the wrist. During the data collection process, the orientation of the gripper was fixed, so that the camera was pointing down along the gripper, with the gripper still being in the field of view. This simplifies the overall setup and keeps the object in view of the camera.

\subsection{Dataset}
\begin{figure*}
\centering
    \begin{subfigure}{0.3\textwidth}
\includegraphics[width=\textwidth]
{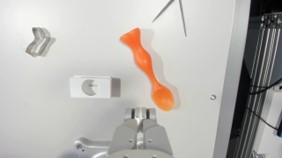}
        \caption{Unusual design (Ice cream former)}
        \label{fig:ice_cream_former}
\end{subfigure}
\begin{subfigure}{0.3\textwidth}
\includegraphics[width=\textwidth]
{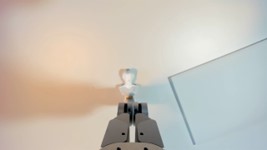}
        \caption{Unusual lighting (Bust toy)}
        \label{fig:bust_toy}
\end{subfigure}
\begin{subfigure}{0.3\textwidth}
\includegraphics[width=\textwidth]
{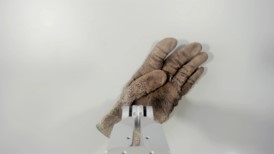}
        \caption{Wide object (Gardening glove)}
        \label{fig:gardening_glove}
\end{subfigure}

\begin{subfigure}{0.3\textwidth}
\includegraphics[width=\textwidth]
{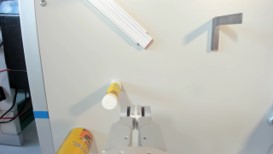
}
        \caption{Vertically shaped object (Glue stick)}
        \label{fig:glue_stick}
\end{subfigure}
\begin{subfigure}{0.3\textwidth}
\includegraphics[width=\textwidth]
{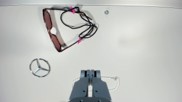
}
        \caption{Irregularly shaped object (Sunglasses)}
        \label{fig:sunglasses}
\end{subfigure}
\caption{Example images for different object categories.}
\end{figure*}

The collected dataset consists of around 750 different objects, with 5 image sequences for each object, taken while the robot moves the Tool Center Point (TCP), which is defined as the middle lowest point of the gripper, from a random upper position to a top center position above the object. The image sequences and metadata are collected synchronized with 2 to 6 frames per second. The dataset is robustified using different lighting situations and by approaching the object not only linearly, but also with a curved and triangular trajectory.
Around 60\% of the sequences contain single and 40\% multi-object settings. Example images from the dataset can be seen in Figs.~ \ref{fig:ice_cream_former}, \ref{fig:bust_toy},
\ref{fig:gardening_glove}, 
\ref{fig:glue_stick}, \ref{fig:sunglasses}.
%%%
%%%
\subsection{Training details}
\begin{figure}[H]
\center
\includegraphics[width=0.45\textwidth]{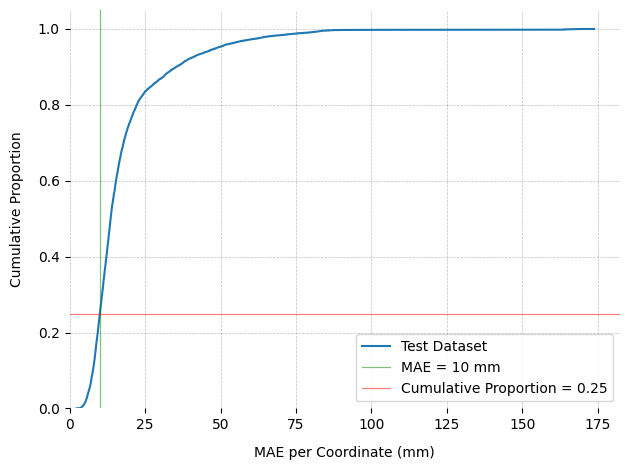}
    \caption{Cumulative Density Function (CDF) plot for error results on testset}
    \label{fig:cdf}
\end{figure}

For training, the data has been split into 90\% training and validation and 10\% testing data. A 5-fold cross-validation was used for each of the tested models.
To prevent leaking of the height of the object, a group-based data split was used for training, validation, and testing, that ensured all images of the same object were in the same set. Huber loss was used for training and validation, Mean Absolute Error (MAE) and Euclidean error were used for testing, and MAE for error evaluation. The MAE was used with the default PyTorch implementation, with a mean reduction over the coordinates. To visually inspect the proportion of samples with an error margin that possibly allows grasping or pushing tasks, a Cumulative Distribution Function (CDF) plot was created, shown in Fig.~\ref{fig:cdf}

 %\begin{figure}[H]
 %    \centering
    % First Figure % Adjust width as needed
 %        \centering
 %        \includegraphics[width=0.3\textwidth]{images/bigger_labels/boxplot_MAE.png}
 %        \caption{Test MAE for LLaVA-v1.5 as base model}
 %        \label{fig:llava_v1.5_mae_errors}
 %\end{figure}

\section{Results}As a baseline, LLaVA-v1.5 was used as a feature extractor with a single linear regression layer. For the task, the model with the 7B LLaVA-v1.5 as a base model showed the best performance and outperformed the baseline by 5 times. The tested models included the 7B Mistral LLaVA-NeXT and LLaVA-Onevision in the 0.5B version. The best model, which uses LLaVA-v1.5 as a base model, plateaus after 10 epochs at around 20, while the baseline shows no real improvements at around 100 Huber loss units on the validation set. For the best model, we performed a detailed error analysis to investigate where and why the model fails. Although our model achieves MAEs below 20mm for 75\%, there are substantially larger errors, above 35mm for around 5\% of the objects.
%, as shown in Fig.~\ref{fig:llava_v1.5_mae_errors}. 
We analyze the outlier counts per object group for all groups with 25 and more outliers, as seen in Table \ref{tab:outlier_counts}.
\begin{table} %[!t]
\center
\begin{footnotesize}
\begin{tabular}{lr}
\toprule
 Group                           &   Count \\
\midrule
 ice cream former                &                293 \\
 water level                     &                274 \\
 manikin hand                    &                234 \\
 toothbrush pack                 &                225 \\
 power supply adapter set        &                218 \\
 pills                           &                217 \\
 soda bottle                     &                207 \\
 spraycan                        &                206 \\
 wasabi                          &                179 \\
 soap dispenser                  &                177 \\
 bottle opener                   &                144 \\
 glue stick                      &                139 \\
 SCALPEL HOLDER,NO.3,12CM        &                139 \\
 bust toy                        &                132 \\
 pack of chopsticks              &                116 \\
 toy candlestick                 &                111 \\
 bandera                         &                110 \\
 sewing kit                      &                 87 \\
 plate                           &                 84 \\
 velvet bag                      &                 82 \\
 projector                       &                 71 \\
 egg tector                      &                 68 \\
 stuffing egg                    &                 66 \\
 screwtop jar                    &                 62 \\
 toy scissors                    &                 57 \\
 14mm wrench                     &                 53 \\
 timer switch                    &                 50 \\
 TCDISSECTING SCISSORS.CVC.,23CM &                 43 \\
 sunglasses                      &                 43 \\
 wallet                          &                 42 \\
 white crayon                    &                 40 \\
 scissors 90degrees              &                 34 \\
 amazon necklace                 &                 28 \\
 bull figurine                   &                 28 \\
 noise stop earplugs             &                 27 \\
 paper cup                       &                 25 \\
%audio cable switch              &                 23 \\
%allen wrench with blue ribbon   &                 19 \\
% ari hand                        &                 18 \\
% hummus                          &                 12 \\
% salt shaker                     &                 11 \\
% key 4                           &                 11 \\
% hook 2                          &                 11 \\
% knitted flower                  &                 10 \\
% AA battery                      &                  9 \\
% energy drink can                &                  8 \\
% dissecting forceps 220mm        &                  8 \\
% bicycle tail light              &                  6 \\
% matchbox                        &                  6 \\
% sewing needle                   &                  6 \\
% hairpin                         &                  5 \\
% straw                           &                  4 \\
% cup 5                           &                  3 \\
% safety pin                      &                  3 \\
% pottery                         &                  1 \\
% cigar                           &                  1 \\
% candy toy                       &                  1 \\
\bottomrule
\end{tabular}
\end{footnotesize}
\caption{Outlier count ($ \text{MAE} > 40 $) per object group}
\label{tab:outlier_counts}
\end{table}
We then formulate hypotheses about the relationships between errors and object properties to understand where the model fails.
The first hypothesis is that some vertically shaped objects, such as "glue stick" and "soda bottle", are harder to predict because they are not properly visible from above. 
The second hypothesis is that unusually designed objects like "ice cream former" are harder to predict because it can be assumed the base model is biased towards conventional design patterns since internet scale training data will overrepresent these. Additional hypotheses are that irregular shapes may be harder to predict, for example "sunglasses", because geometry helps finding the top center point, and that positions of wider objects are harder to predict, such as "gardening glove", because the object might not be fully visible the closer the camera gets. Finally, it is assumed that the z-coordinate is harder to predict than the other coordinates in this setting, since predicting object height is challenging with a monocular RGB image only. 
Image examples of objects in the test set in relation to the hypotheses can be seen in Figs.~ \ref{fig:ice_cream_former}, \ref{fig:bust_toy},
\ref{fig:gardening_glove}, 
\ref{fig:glue_stick}, \ref{fig:sunglasses}.

\begin{figure*}
    \centering
\begin{subfigure}{0.32\textwidth}
\includegraphics[width=\textwidth]{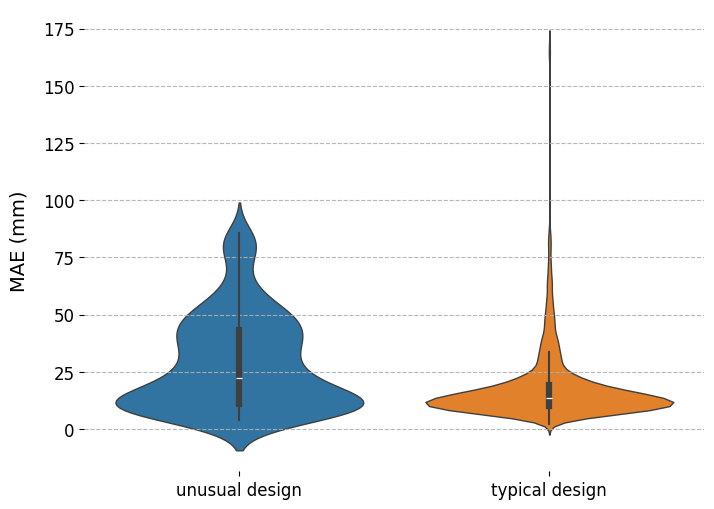}
        \caption{Design patterns}
        \label{fig:unusual_design}
\end{subfigure}
    \begin{subfigure}{0.32\textwidth}
\includegraphics[width=\textwidth]{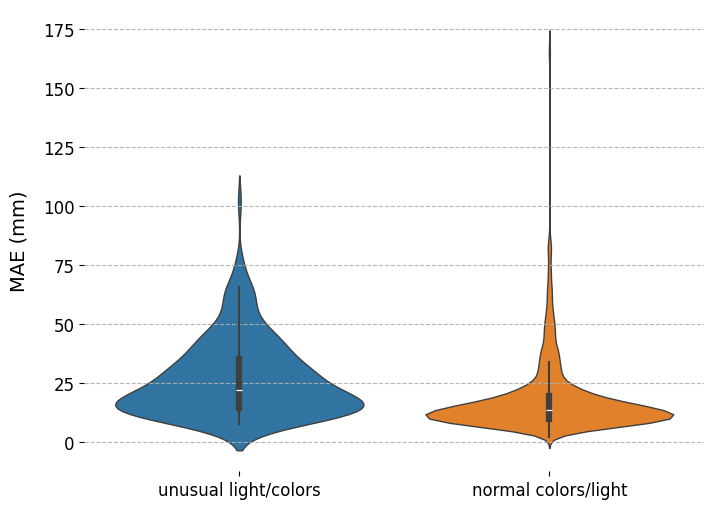}
        \caption{Lighting and color in image}
        \label{fig:unusual_light}
\end{subfigure}
    \begin{subfigure}{0.32\textwidth}
\includegraphics[width=\textwidth]{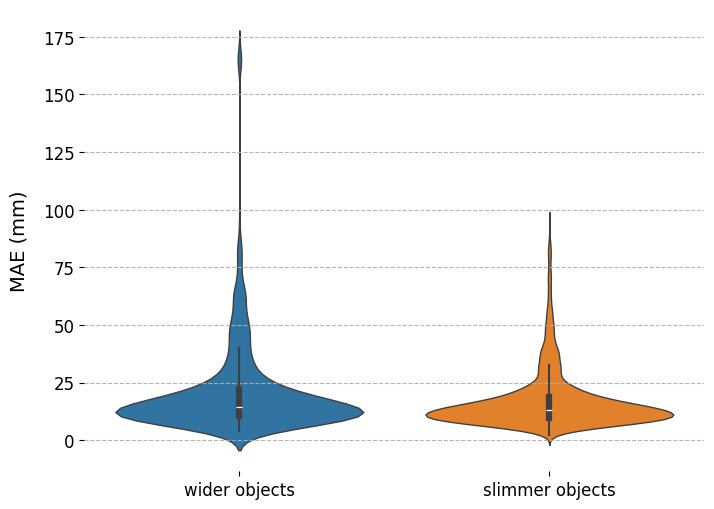}
        \caption{Relative object size in image}
        \label{fig:wider_objects}
\end{subfigure}
    \begin{subfigure}{0.32\textwidth}
\includegraphics[width=\textwidth]
{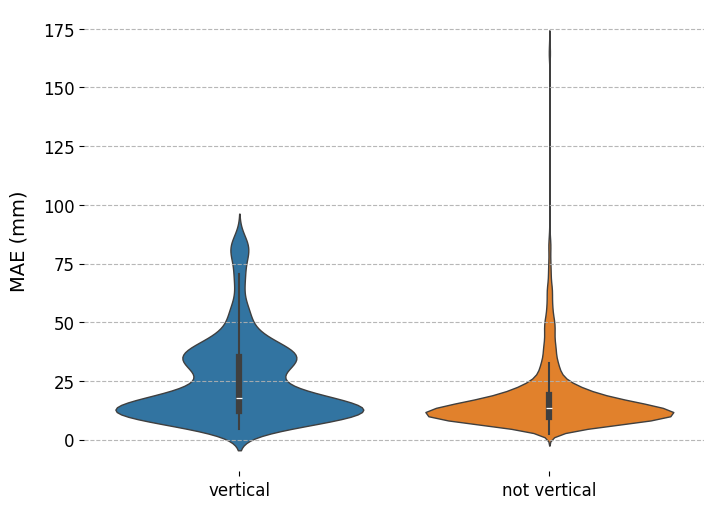}
        \caption{Object verticality}
        \label{fig:vertically_shaped}
\end{subfigure}
    \begin{subfigure}{0.32\textwidth}
\includegraphics[width=\textwidth]{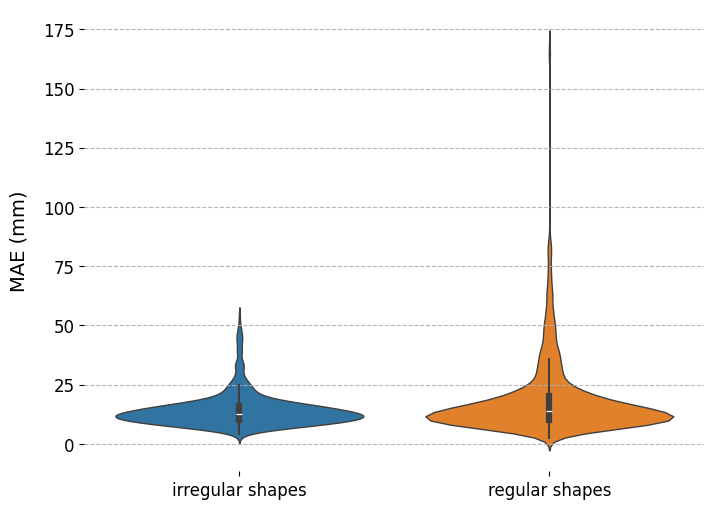}
        \caption{Shapes regularity}
        \label{fig:regular_shapes}
\end{subfigure}
\caption{Error distributions for different object and image properties.}
\end{figure*}
\begin{figure} %[!t]
    %\ContinuedFloat
    \centering
\includegraphics[width=0.38\textwidth]
{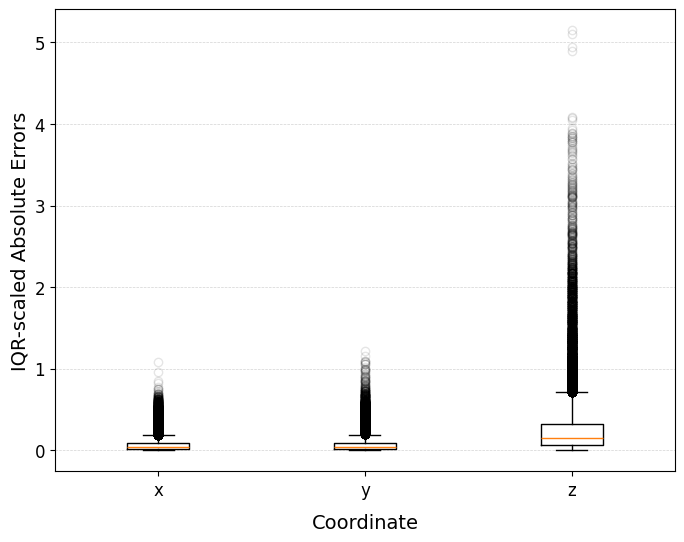}
    \caption{Comparison of IQR-scaled absolute errors per coordinate}
    \label{fig:IQR_scaled_errors}
\end{figure}

\section{Discussion}
Hypothesis testing revealed that most of the hypotheses were confirmed. Violin plots of the distribution of the respective groups are shown in Fig.~\ref{fig:vertically_shaped}, \ref{fig:regular_shapes}, \ref{fig:unusual_design}, \ref{fig:wider_objects}, \ref{fig:unusual_light}. Samples with vertically shaped objects compared to those without vertically shaped objects were shown to have higher prediction errors on the test set. In addition, unusually designed objects,
as well as wider objects and objects with unusual light or color conditions in the images, had higher prediction errors. This was also confirmed with Mann-Whitney U tests, which showed that the difference between the two groups is significant to a significance level of 0.05 with corrected p-values after Bonferroni correction.
The prediction of the z-coordinate, which signifies the height of an object, has errors that are relatively higher than the x and y coordinates. This could be observed after Inter-Quartile-Range(IQR)-scaling the errors, which enables fair comparison across different range. The x, y coordinates show similar distributions, while z-coordinate errors exhibit larger spread, indicating higher uncertainty in depth estimation from monocular images (see Fig.~\ref{fig:IQR_scaled_errors}). 
The model generalized well on unseen data during validation and testing on a dataset which includes a wide range of object shapes, light situations, and object positions. Our best model's test performance should be sufficient for most cases. Since the median Euclidean error is around 27 mm and the median Absolute Error with mean reduction is around 13 mm, which we consider this a success for an open-set prediction task. From the CDF plot for the results on the test set, it can be seen that 25 \% of the samples are within a mean 10 mm error per coordinate, which implies potentially successful trials for tasks such as grasping or pushing.

\section{Conclusion and Outlook}
The model generalizes well on different objects for open set prediction. But the model is still biased towards the workspace of the robot arm, including the camera's model, since the model, has only been trained on the specifically collected dataset. The future goal is to enhance the dataset's heterogeneity by incorporating more multi-object settings and increasing variations in how the object is approached and ideally also include different workspaces and robot models. Improvements cannot only be made for the training data but also on the model side.  We intend to simplify the prompt input and introduce the robot proprioceptive data during late feature fusion. Future work will also adopt learned routing strategies. 

%\newpage
%\clearpage
%\input{appendix}

\bibliographystyle{IEEEbib}
\bibliography{refs_shortAuthorLists}
\end{document}